\documentclass[11pt]{article}

\usepackage[utf8]{inputenc}
\usepackage[T1]{fontenc}
\usepackage{amsmath,amssymb,amsthm}
\usepackage{graphicx}
\usepackage{booktabs}
\usepackage[hypertexnames=false]{hyperref}
\usepackage{url}
\usepackage{xspace}
\usepackage{microtype}
\usepackage{listings}
\usepackage{xcolor}
\usepackage{algorithm}
\usepackage{algpseudocode}
\usepackage[margin=1in]{geometry}

\hypersetup{
    colorlinks=true,
    linkcolor=blue,
    filecolor=magenta,
    urlcolor=cyan,
    citecolor=blue,
    pdftitle={Binary BPE: A Family of Cross-Platform Tokenizers for Binary Analysis},
    pdfauthor={Michael J. Bommarito II},
    pdfsubject={Binary Analysis, Machine Learning, Tokenization},
    pdfkeywords={binary analysis, tokenization, byte pair encoding, malware detection, machine learning}
}

\newcommand{\tokenizername}{Binary BPE\xspace}
\newcommand{\tokenizerfamilyname}{Binary BPE family\xspace}

\title{Binary BPE: \\ A Family of Cross-Platform Tokenizers for Binary Analysis}

\author{
Michael J. Bommarito II\thanks{Portions of this work were prepared with assistance from large language models. The author is solely responsible for all content, including any errors or omissions.} \\
\texttt{michael.bommarito@gmail.com}
}

\date{November 2025}

\begin{document}

\maketitle

\begin{abstract}
Sequence models for binary analysis are bottlenecked by byte-level tokenization: raw bytes waste precious context window capacity for transformers and other neural network architectures, and many existing text-oriented tokenizers fail on arbitrary 0x00--0xFF sequences. To address this issue, we introduce the \tokenizerfamilyname, a set of cross-platform Byte Pair Encoding (BPE) tokenizers for executables trained on a large corpus of binaries spanning multiple platforms, architectures, and operating systems, including Linux, Windows, macOS, Android, and malware sources. We release trained tokenizers with vocabularies of 4K, 8K, 16K, 32K, and 64K tokens, enabling both systematic scaling studies and practical deployment from resource-constrained edge devices to high-throughput datacenters. The family exhibits a nested hierarchy where each smaller vocabulary is a perfect prefix of larger ones, enabling seamless transfer of learned embeddings when scaling model capacity. These tokenizers discover interpretable patterns (ELF/PE headers, instruction sequences, cross-platform strings) while yielding multi-byte compression per token. On representative \emph{uncompressed} executables (e.g., ELF/PE/Mach-O rather than compressed APKs), the \tokenizerfamilyname{} typically allows for roughly 2--3$\times$ more binary content per fixed-length transformer context window than raw bytes, enabling more efficient research and practical deployment for content identification, malware detection, reverse engineering, and optimization. We release the trained Binary BPE tokenizers on HuggingFace and supporting code on GitHub, providing a drop-in, open-source foundation for binary-focused language models and context-efficient agentic tools.
\end{abstract}

\section{Introduction}
\label{sec:introduction}

Sequence models---including recurrent neural networks (RNNs), convolutional neural networks (CNNs), structured state-space models (SSMs), and most especially transformers---require tokenization to convert input data into discrete sequences. Existing tokenizers like BPE~\cite{sennrich2016neural}, WordPiece~\cite{schuster2012japanese}, and SentencePiece~\cite{kudo2018sentencepiece} are designed for natural language and source code, typically learning subword units that respect linguistic structure, word boundaries, and UTF-8 text properties. Binary executables, however, present a fundamentally different domain: raw byte sequences (0x00--0xFF) containing machine code instructions, file format headers, compressed resources, and encrypted sections with no linguistic regularities, word boundaries, or printable characters. Applying NLP tokenizers to binaries fails catastrophically, as they are optimized for human-readable text, not arbitrary byte distributions. Processing binaries as raw byte sequences also wastes tokens: a 100KB binary encoded byte by byte produces 100K tokens, quickly exhausting typical accelerator memory, especially when malware and production binaries are often orders of magnitude larger.

Developing effective binary tokenization faces several unique challenges. First, \emph{platform diversity}: Linux (ELF), Windows (PE), macOS (Mach-O), and Android (APK) employ distinct file format conventions with different header structures and section layouts. Second, \emph{architectural heterogeneity}: x86-64, ARM, MIPS, and RISC-V instruction sets encode operations differently, creating ISA-specific byte patterns that must be learned jointly for cross-platform utility. Third, \emph{extreme length variation}: binaries range from 10KB utilities to multi-megabyte applications, with sections (code, data, resources) exhibiting dramatically different compression characteristics. Finally, \emph{domain variability}: system utilities, GUI applications, and malware samples contain distinct byte-level distributions. A successful tokenizer must learn patterns that generalize across these dimensions while preserving format- and ISA-specific structure.

Binary analysis underpins tasks such as malware detection and classification, vulnerability discovery in proprietary software, and program understanding or reverse engineering of stripped or obfuscated executables. Recent advances demonstrate that multiple hybrid or transformer-based architectures with scaled attention can learn rich binary representations for these tasks, but their effectiveness is fundamentally constrained by tokenization. While tokenization critically impacts transformer performance on code~\cite{mostafa2025tokenization}, existing approaches operate on disassembled text, limiting applicability to stripped or obfuscated binaries. Android-specific methods~\cite{rahali2023malbertv2} apply BPE to decompiled manifest files but require high-level symbolic information. Raw byte transformers~\cite{lu2025progressive} avoid disassembly but lack learned vocabulary compression. The challenge remains: can one develop a byte-level BPE tokenizer that learns cross-platform patterns from raw binaries without disassembly or decompilation?

Motivated by these opportunities, we present a \emph{family} of cross-platform BPE tokenizers for raw binary executables—infrastructure enabling transformer-based and more broadly sequence-model-based binary analysis. Trained on 24~GB of data across 30,000 unique binaries from the Binary-30K dataset~\cite{bommarito2025binary30k}, the family achieves roughly 2--3$\times$ effective compression on typical \emph{uncompressed} binaries (ELF/PE/Mach-O), while compressed, high-entropy formats such as Android APKs naturally see smaller but still meaningful gains. Beyond compression, BPE tokenization also performs unsupervised pattern discovery that pre-encodes domain structure before transformer training begins. By learning frequent byte co-occurrences, BPE discovers and consolidates recurring patterns—x86-64 instruction prefixes (REX.W), file format headers (ELF magic), cross-platform string fragments—into single vocabulary tokens. This gives transformers meaningful semantic units rather than forcing models to rediscover "these bytes always co-occur" from scratch during training. For binary analysis, where ISA-specific encodings and format conventions create predictable byte-level structure, BPE effectively performs feature learning at the byte level, enabling more efficient downstream representation learning.

The vocabulary family serves dual purposes. \emph{Scientifically}, it enables controlled ablation studies of BPE scaling behavior on binary data, quantifying compression-vocabulary tradeoffs, sequence coverage dynamics, and learned pattern evolution across architectures—questions otherwise unanswerable with a single model. \emph{Practically}, different deployment scenarios demand different resource-computation tradeoffs. Smaller vocabularies (4K, 8K) enable deployment in resource-constrained environments like network security appliances performing inline packet inspection, edge devices with limited memory footprints, or embedded malware scanners in gateways.  Where tokenizer size and inference latency are critical constraints, these smaller vocabulary sizes provide an ideal footprint. Larger vocabularies (32K, 64K) maximize compression efficiency for datacenter-scale analysis, where accelerator memory and throughput dominate. This range supports both research and pedagogy on tokenization fundamentals, as well as real-world deployment from edge to cloud. Critically, the family exhibits a nested hierarchy: each smaller vocabulary is a perfect prefix of all larger ones, with identical token IDs mapping to identical byte sequences. This property enables seamless model scaling—practitioners can train a 1B-parameter model with 4K embeddings, then expand to a 7B-parameter model with 64K embeddings while preserving all originally learned representations, avoiding costly retraining of the base vocabulary. Beyond serving as a drop-in tokenizer for larger general-purpose models, the \tokenizerfamilyname{} is designed to support smaller specialist masked- and causal-language models that might act as tools inside larger analysis workflows; for example, plug-ins embedded in reverse-engineering frameworks, such as Ghidra or IDA Pro, or domain-specific binary ``experts'' orchestrated by agentic systems or human analysts.

In this paper, we contribute: (1)~a \textbf{training methodology} for BPE on diverse binaries (ELF, PE, Mach-O, APK) across x86-64, ARM, MIPS, and RISC-V architectures, with platform stratification to ensure cross-platform pattern learning; (2)~an \textbf{ablation study} of vocabulary size (4K, 8K, 16K, 32K, 64K) quantifying compression, coverage, and token-length shifts; (3)~ \textbf{analysis of learned patterns} demonstrating instruction- and format-aware tokens across sizes; and (4)~public release of the \tokenizerfamilyname{} on HuggingFace, together with replication for our publication experiments, figures, and tables.

\section{Related Work}
\label{sec:related}

\subsection{Tokenization in Natural Language Processing}

Subword tokenization through BPE~\cite{sennrich2016neural}, WordPiece~\cite{schuster2012japanese}, and SentencePiece~\cite{kudo2018sentencepiece} is standard for neural language models. These methods typically assume UTF-8 text with linguistic structure, integrate pre- and post-processing for word boundaries and character composition rules, and feature vocabulary sizes on the order of tens to hundreds of thousands of tokens to cover large lexicons~\cite{devlin2019bert,wolf2020transformers}. Binaries, on the other hand, contain arbitrary byte sequences (0x00--0xFF) with no linguistic regularities, requiring tokenization methods that have no initial assumptions about input structure.

\subsection{Machine Learning for Binary Analysis}

Machine learning for binary analysis falls into three categories. \emph{Raw byte methods}~\cite{raff2018malconv, lu2025progressive} use neural networks like CNNs or transformers with byte-level embeddings, wasting context on individual bytes rather than learning semantic units. For example, processing a 100KB binary byte-by-byte requires 100K tokens, consuming a large amount of accelerator memory before materializing the rest of the architecture. \emph{Engineered feature methods}~\cite{anderson2018ember, liu2024assemblage} extract import tables, section headers, and string statistics, enabling efficient classification but sacrificing structural information for generative tasks and requiring extensive format and architecture coverage and preprocessing. \emph{Assembly-based methods}~\cite{shin2015recognizing, xu2017neural} apply NLP to disassembled instruction sequences but again introduce architecture dependencies and fail on obfuscated code, data sections, or many edge cases.

\subsection{Tokenization for Binary Data}

Systematic tokenization of raw executable bytes remains largely unexplored. Recent work~\cite{mostafa2025tokenization} compares BPE, WordPiece, and Unigram on disassembled C functions but operates on post-disassembly text, requiring decompilation and failing on stripped or obfuscated binaries. Android-specific methods~\cite{rahali2023malbertv2} apply BPE to decompiled manifest XML, requiring symbolic metadata unavailable in stripped native binaries. Fixed n-gram features~\cite{raff2018malconv} lack adaptive vocabulary learning. Work on function embeddings~\cite{massarelli2019safe, arakelyan2021bin2vec} does not address whole-file tokenization. Recent notable industry systems~\cite{stortz2025crowdstrike} use byte-level transformers with convolutional compression but lack public documentation or release, limiting reproducibility. Prior to this work, we are not aware of any openly available cross-platform BPE tokenizer that operates directly on raw executable bytes across ELF, PE, Mach-O, and APK formats.

\subsection{Gap and Contribution}

Existing methods cannot handle binary executables' extreme lengths (100 KB–100 MB), multi-platform diversity (Linux, Windows, macOS, Android), architecture heterogeneity (x86, ARM, MIPS, RISC-V), and format complexity. While industry systems~\cite{stortz2025crowdstrike} may use learned tokenization internally, these models are proprietary and not available to researchers. This lack of public resources prevents reproducible research and forces academics to either process raw bytes inefficiently or build custom tokenizers from scratch for each study.

We introduce the \tokenizerfamilyname, a set of openly available BPE tokenizers trained directly on raw executable bytes across platforms and architectures. The family includes models at 4K, 8K, 16K, 32K, and 64K vocabulary sizes, enabling both systematic scaling studies and practical deployment. The Binary BPE tokenizers themselves are released as HuggingFace-compatible models, with the underlying \texttt{bbpe} Rust trainer also open-sourced for researchers who wish to retrain or extend the vocabulary~\cite{bbpe2025,binarybpe2025}.

\section{Methodology}
\label{sec:methodology}

We design our training methodology around three key components: (1)~stratified corpus construction for cross-platform representation; (2)~vocabulary configuration balancing expressiveness and efficiency; and (3)~evaluation metrics validating both quantitative compression and qualitative pattern learning.

\subsection{Training Data Selection}

We train the \tokenizerfamilyname{} on a curated corpus of approximately 30{,}000 binary records totaling 24~GB.  This corpus includes system utilities, shared libraries, drivers, applications, and malware samples, and is a superset of the Binary-30K dataset~\cite{bommarito2025binary30k}.  Our companion paper on Binary-30K provides a full description of corpus construction, deduplication, labeling, and licensing; here we summarize the aspects most relevant to tokenizer behavior.

The training corpus spans Linux (Alpine, Debian, Ubuntu, BusyBox), Windows (8/10/11), macOS (x86-64, ARM64, and Universal binaries), Android APKs, and malware drawn from SOREL-20M~\cite{harang2020sorel} and Malware Bazaar~\cite{malwarebazaar}. Architecture coverage includes x86-64, x86-32, ARM64, ARM32, MIPS, and RISC-V, and file formats include ELF, PE, Mach-O, and APK.

We stratify sampling by platform and file type to prevent any single operating system from dominating the learned vocabulary and overwhelming cross-platform sequences such as null padding, ASCII strings, and instruction prefixes. Preliminary Linux-only experiments produced noticeably worse compression on Windows and macOS binaries, empirically confirming the need for platform-balanced training when learning a shared cross-platform tokenizer.

\subsection{BPE Training Process}

We apply Byte Pair Encoding (BPE)~\cite{gage1994bpe,sennrich2016neural} directly to raw binary data, treating each executable as a sequence over the complete 256-byte alphabet (0x00--0xFF) with no linguistic assumptions. Starting from this base alphabet, training iteratively merges the most frequent adjacent byte pair until reaching the target vocabulary size, computing merge statistics globally across the cross-platform corpus.

Our \texttt{bbpe} training implementation~\cite{bbpe2025} supports chunk-based processing with entropy filtering to improve merge quality. For all released tokenizers, we use an 8KB chunk size with a maximum entropy threshold of 7.0 bits per byte, filtering out high-entropy blocks that likely contain compressed, encrypted, or random data. This filtering prevents noisy frequency statistics from such regions, which offer no learnable structure, from degrading the quality of proposed merges and ensures the vocabulary focuses on compressible byte patterns present in executable code and data sections.

\textbf{Vocabulary size and implementation.} For our largest tokenizer, we target 65,536 total token IDs (2$^{16}$) to balance expressiveness and efficiency: 65,529 learned BPE tokens (256 base-byte symbols plus 65,273 merges) plus 7 special control tokens (\texttt{<|start|>}, \texttt{<|end|>}, \texttt{<|pad|>}, \texttt{<|unk|>}, \texttt{<|cls|>}, \texttt{<|sep|>}, \texttt{<|mask|>}). Power-of-2 sizing enables compact 16-bit token IDs and keeps the vocabulary comparable to modern LLM/VLM embedding sizes ~\cite{devlin2019bert,wolf2020transformers}. Special tokens are added only after training and never participate in merges, so they support model control without altering the learned byte-level distribution. Unless otherwise noted, structural statistics in Section~\ref{sec:results} are reported over the 65,529 learned BPE tokens, with the 7 special control tokens excluded from aggregate counts. We implement this procedure using a custom Rust BPE trainer (\texttt{bbpe})~\cite{bbpe2025} that emits HuggingFace-compatible tokenizer JSONs~\cite{huggingface2019tokenizers}; the released \tokenizerfamilyname{} models (\texttt{mjbommar/binary-tokenizer-001-\{4k,8k,16k,32k,64k\}}) are published on HuggingFace, and the \texttt{bbpe} trainer is available in the Binary BPE repository for practitioners who need to retrain or adapt the vocabulary~\cite{binarybpe2025}.

\paragraph{Family training setup.} To study scaling, we train tokenizers at 4K, 8K, 16K, 32K, and 64K vocabulary sizes under the same corpus and hyperparameters, varying only the target vocabulary size. By design, BPE training produces a nested hierarchy: each smaller vocabulary is a perfect prefix of all larger ones. Token ID 0--4095 in the 4K tokenizer map to identical byte sequences in all larger tokenizers, and similarly for 8K, 16K, and 32K. This deterministic nesting enables embedding transfer when scaling model capacity—practitioners can initialize a larger model's embedding matrix with weights from a smaller model, preserving learned representations for the shared vocabulary while learning only the new tokens. All results below compare models trained with this unified protocol.

\subsection{Evaluation Metrics}

We assess tokenizer quality along four complementary dimensions:

\textbf{1. Vocabulary structure} uses the released tokenizer JSONs to compute token length distributions, coverage of short byte sequences, and mutually exclusive content categories (base alphabet, readable strings, instruction patterns, padding, high-byte-only tokens, and mixed structures). These metrics depend only on the vocabulary and merges, not on any downstream dataset.

\textbf{2. Single-file compression} measures bytes per token on representative binaries that are widely available on most systems. In particular, we report detailed results for \texttt{/usr/bin/ls}, which serves as a simple, reproducible benchmark for vocabulary-size scaling across the \tokenizerfamilyname.

\textbf{3. Stratified small-benchmark compression} evaluates the 64K tokenizer on 24 representative binaries covering multiple operating systems, architectures, file formats, and both benign and malicious samples. This benchmark enables lightweight reproduction without requiring the full Binary-30K dataset.

\textbf{4. Corpus summary statistics} provide population-level statistics for our training corpus. We summarize platform-level averages like tokenized length, unique-token counts, and bytes per token from the full Binary-30K dataset~\cite{bommarito2025binary30k} using the 64K tokenizer. The Binary-30K paper remains the authoritative source for corpus construction, labeling, and licensing; here, we use only aggregate metrics derived from its tokenized form.

These metrics enable comprehensive evaluation: vocabulary structure and scaling experiments reveal what patterns the tokenizer learns; the \texttt{/usr/bin/ls} and \texttt{samples/} benchmarks quantify practical compression on concrete binaries; and large-corpus statistics verify that these behaviors persist on a substantially larger dataset without requiring full dataset access for reproduction.

\section{Results}
\label{sec:results}

We evaluate \tokenizername{} along three dimensions: vocabulary structure (Section~\ref{sec:methodology}), compression behavior on concrete binaries, and scaling across vocabulary sizes. The first set of results is derived directly from the released tokenizer JSONs and characterizes token lengths and content categories. The second set uses fully reproducible benchmarks---\texttt{/usr/bin/ls} and the stratified \texttt{samples/} directory---to measure bytes per token on real binaries. Finally, we summarize full corpus statistics from the Binary-30K dataset~\cite{bommarito2025binary30k} and study vocabulary-size scaling across the \tokenizerfamilyname{} in Section~\ref{sec:scaling}. Together, these results demonstrate that BPE successfully learns cross-platform binary structures without supervision while providing practical compression for modern sequence models.

\subsection{Vocabulary Analysis}

The 64K tokenizer vocabulary consists of 65,536 tokens in total—65,529 learned BPE tokens plus 7 special tokens—and exhibits rich structural diversity. Table~\ref{tab:token_lengths} shows the complete length distribution over the learned BPE tokens (special control tokens are excluded from these counts), with token lengths ranging from 1 byte to 32 bytes and a median length of 3 bytes (mean: 4.17 bytes). Two-byte tokens dominate the vocabulary, capturing frequent instruction prefixes and byte pairs, and 3-byte tokens align closely with common x86-64 instruction encodings (REX prefix + opcode + ModR/M), demonstrating architecture-aware learning without explicit programming. Longer patterns encode format headers, extended instructions, and string fragments, while very long tokens represent complete format structures or long strings.

\begin{table}[t]
\centering
\begin{tabular}{lrrl}
\toprule
\textbf{Length (bytes)} & \textbf{Count} & \textbf{\%} & \textbf{Pattern Type} \\
\midrule
1 & 256 & 0.4 & Base alphabet \\
2 & 24,944 & 38.1 & Instruction prefixes \\
\textbf{3} & \textbf{11,730} & \textbf{17.9} & x86-64 instructions \\
4 & 13,189 & 20.1 & Format headers \\
5--8 & 10,908 & 16.6 & Medium patterns \\
9--16 & 3,420 & 5.2 & Long patterns \\
17--32 & 1,082 & 1.7 & Very long patterns \\
\midrule
Total & 65,529 & 100.0 & Median: 3, Mean: 4.17 \\
\bottomrule
\end{tabular}
\caption{Token length distribution for the \tokenizername{} learned BPE vocabulary (65{,}529 tokens; 7 special control tokens are excluded from these counts). Median length is 3 bytes, aligning with common x86-64 instruction encoding (REX prefix + opcode + ModR/M byte). Two-byte tokens dominate, representing frequent instruction prefixes and byte pairs.}
\label{tab:token_lengths}
\end{table}

Table~\ref{tab:vocab_analysis} categorizes the vocabulary into six mutually exclusive types. The base alphabet provides complete byte coverage, readable string tokens capture file paths and identifiers, and more than a quarter of the vocabulary consists of instruction-pattern tokens (e.g., x86-64 REX prefixes and common opcodes). Additional categories include high-byte-only tokens, which represent compressed or encrypted regions, and mixed tokens that span format headers, operands, and embedded strings. Together, these distributions show that the tokenizer allocates substantial capacity to both human-readable strings and CPU instructions.

\begin{table}[t]
\centering
\begin{tabular}{lrrl}
\toprule
\textbf{Category} & \textbf{Count} & \textbf{\%} & \textbf{Description} \\
\midrule
Base alphabet & 256 & 0.4 & Single-byte (0x00--0xFF) \\
Pure null padding & 31 & 0.0 & All nulls (alignment) \\
Readable strings & 8,211 & 12.5 & Letters, digits, symbols \\
Instruction patterns & 18,013 & 27.5 & x86/ARM opcodes \\
High bytes only & 4,920 & 7.5 & Non-ASCII ($\geq$0x80) \\
Other & 34,105 & 52.0 & Mixed/format structures \\
\bottomrule
\end{tabular}
\caption{Vocabulary composition of the \tokenizername{} (65{,}536 tokens total: 65{,}529 learned BPE tokens plus 7 special control tokens). Counts and percentages are computed over the full 65{,}536-token vocabulary, with special control tokens included in the appropriate content categories. Categories are mutually exclusive and ordered by priority: base alphabet, readable strings, instruction patterns, padding, high bytes, and other mixed content.}
\label{tab:vocab_analysis}
\end{table}

Beyond these structural statistics, the vocabulary itself reveals distinct pattern categories that explain the tokenizer's cross-platform effectiveness.

\subsection{Learned Patterns and Interpretation}

Without explicit supervision, the tokenizer discovers three pattern categories that explain its cross-platform effectiveness. Critically, BPE on binary data is not merely compression; it performs unsupervised feature discovery, surfacing hierarchical structures that enable transformers to operate at the granularity of instructions and format components rather than individual bytes.

\textbf{File format structures} provide format-level abstraction. Magic numbers (\texttt{\textbackslash x7fELF}, \texttt{MZ}, \texttt{\textbackslash xFE\textbackslash xED\textbackslash xFA\textbackslash xCE}) emerge as single tokens despite appearing only at file starts; BPE's frequency-based merging discovers invariant structures even with low absolute frequency. ELF section names (\texttt{.text}, \texttt{.rodata}), PE import tables, and Mach-O load commands similarly merge. These tokens allow downstream transformers to immediately recognize file types and structural boundaries without learning byte-level compositions—analogous to how NLP tokenizers surface morphemes or structural formatting tokens like \texttt{<br>} in HTML or \texttt{\#\#} in Markdown.

\textbf{Architecture-specific patterns} capture instruction set regularities. x86-64 REX.W prefix (\texttt{\textbackslash x48}) with MOV opcodes form frequent 2-3 byte tokens. The tokenizer learns 11,730 length-3 tokens (17.9\% of vocabulary), closely matching common x86-64 instruction length (REX + opcode + ModR/M)—an alignment discovered through corpus statistics, not engineering. Direct vocabulary inspection reveals 18,013 tokens (27.5\%) containing x86 instruction patterns (REX prefixes 0x48--0x4F, or common opcodes for MOV, CALL, JMP, PUSH, POP), confirming unsupervised discovery of ISA-specific encodings. This means more than a quarter of the vocabulary is dedicated to representing CPU instructions as atomic units.

\textbf{Cross-platform byte sequences} reflect OS-level conventions. Null byte padding (4-, 8-, 16-byte runs for alignment), ASCII library paths (\texttt{/lib/}, \texttt{/usr/lib/}), and dynamic linking strings (\texttt{.so}, \texttt{.dll}, \texttt{.dylib}) tokenize consistently across platforms because these conventions are independent of file format or architecture.

The learned vocabulary serves as an intermediate representation between raw bytes and high-level semantics, addressing the challenge of handling multi-platform, multi-architecture diversity without manual feature engineering.

These learned patterns translate to practical performance on real binaries.

\subsection{Compression Efficiency}

To measure compression effectiveness, we evaluate the 64K tokenizer on two benchmarks: the stratified 24-file \texttt{samples/} set (Section~\ref{sec:methodology}) and widely available system binaries such as \texttt{/usr/bin/ls}.

Table~\ref{tab:samples_compression} summarizes bytes per token on the \texttt{samples/} benchmark. Across the 24 binaries, the tokenizer achieves an overall mean of 2.62~bytes/token, with benign and malicious binaries across platforms falling in a narrow band around this value. Intuitively, each token represents a few bytes of input on realistic mixed workloads, rather than a single byte.

These ratios translate directly into context-window efficiency. At 2.6~bytes/token, an 8{,}192-token context can cover about 21~KB of binary content, and a 32{,}768-token context can cover roughly 84~KB—often enough to include complete small binaries or large slices of complex executables in a single forward pass.

As a large-corpus sanity check, Table~\ref{tab:large_corpus_stats} reports aggregate statistics from a precomputed JSON summary file that summarizes the Binary-30K dataset tokenized with the 64K tokenizer. Bytes per token remain in the same 2--3.5~B/token range across platforms, and typical binaries span tens of thousands to millions of tokens, reinforcing the need for compression and motivating architectures that can exploit long contexts. We rely only on these aggregate statistics; the Binary-30K paper provides the detailed description of the underlying corpus.

Taken together, the vocabulary analysis and compression benchmarks show that BPE on raw binaries yields interpretable tokens while achieving multi-byte compression per token across benign and malicious binaries. Section~\ref{sec:scaling} analyzes how these properties evolve as we vary vocabulary size within the \tokenizerfamilyname{}, using \texttt{/usr/bin/ls} and additional mixed-file benchmarks to study 4K--64K configurations.

\begin{table}[t]
\centering
\begin{tabular}{lrr}
\toprule
\textbf{Group} & \textbf{Bytes/Token} & \textbf{$n$}\\
\midrule
Linux & 2.571 & 11\\
Windows & 3.385 & 3\\
Malware (android apk) & 1.411 & 1\\
Malware (linux elf) & 1.834 & 4\\
Malware (macos mach o) & 3.058 & 2\\
Malware (windows pe) & 3.164 & 3\\
\midrule
\textbf{Overall} & \textbf{2.616} & \textbf{24}\\
\bottomrule
\end{tabular}
\caption{Bytes per token for the Binary BPE 64K tokenizer on the stratified \texttt{samples/} benchmark set (24 binaries after extracting \texttt{samples.zip}). Values report mean bytes per token per group; $n$ is the number of binaries.}
\label{tab:samples_compression}
\end{table}

\begin{table}[t]
\centering
\begin{tabular}{lrrrr}
\toprule
\textbf{Platform} & \textbf{Avg Tokens} & \textbf{Median Tokens} & \textbf{Avg Unique} & \textbf{Bytes/Token}\\
\midrule
Linux & 76,880 & 19,861 & 6,134 & 3.339\\
Windows & 258,886 & 49,823 & 12,850 & 3.484\\
macOS & 384,649 & 336,476 & 16,147 & 2.836\\
Android & 2,445,585 & 1,454,293 & 28,133 & 1.996\\
Other & 71,770 & 2,686 & 1,511 & 3.048\\
\bottomrule
\end{tabular}
\caption{Large-corpus summary for the Binary BPE 64K tokenizer on the Binary-30K dataset (aggregate statistics from \texttt{tokenization\_stats.json}). Values report average and median tokenized lengths, average number of unique tokens, and mean bytes per token per platform.}
\label{tab:large_corpus_stats}
\end{table}

\section{Vocabulary-Size Scaling Ablation}
\label{sec:scaling}

For our ablations, we trained the \tokenizerfamilyname{} with vocabularies of 4{,}096, 8{,}192, 16{,}384, 32{,}768, and 65{,}536 tokens. All tokenizers share the same corpus and training protocol; Section~\ref{sec:methodology} details the dataset and procedure. Table~\ref{tab:scaling_summary} summarizes core vocabulary statistics derived from the released models on Hugging Face.

\begin{table}[t]
\centering
\begin{tabular}{lrrrrr}
\toprule
\textbf{Vocab} & \textbf{Avg len} & \textbf{2B cov.} & \textbf{3B share} & \textbf{High-byte} & \textbf{ASCII-only} \\
\midrule
  4K  & 3.000 & 3.01\% & 20.57\% & 54.0\% & 46.0\% \\
  8K  & 3.312 & 5.58\% & 21.67\% & 53.5\% & 46.5\% \\
 16K  & 3.498 & 10.91\% & 20.52\% & 55.7\% & 44.3\% \\
 32K  & 3.812 & 20.49\% & 19.47\% & 57.9\% & 42.1\% \\
 64K  & 4.173 & 38.06\% & 17.90\% & 62.3\% & 37.7\% \\
\bottomrule
\end{tabular}
\caption{Binary BPE tokenizer family scaling summary from released tokenizer JSONs (65{,}536 tokens total per model, including 7 special tokens). Coverage is fraction of all 2-byte sequences; ``3B share'' is percent of vocabulary with length 3.}
\label{tab:scaling_summary}
\end{table}

\paragraph{Trends.} Scaling vocabulary size steadily increases average token length (from 3.00 to 4.17 bytes between 4K and 64K) and expands coverage of 2-byte sequences (from 3.0\% to 38.1\%). Coverage of 3-byte sequences remains low across all vocabulary sizes. The proportion of 3-byte tokens---aligned with common x86-64 instruction length (REX + opcode + ModR/M)---peaks near 8K before giving way to longer merges, and content gradually shifts toward high-byte-only tokens, indicating richer binary-only patterns (less pure ASCII-only content). We validate these structural changes through compression benchmarks on representative binaries.

\paragraph{Compression vs. vocabulary.} As anecdotal evidence from a representative Linux binary (\texttt{/usr/bin/ls}), bytes per token increase monotonically with vocabulary size, from 1.77~bytes/token at 4K to 2.53~bytes/token at 64K. Table~\ref{tab:ls_visual_ablation} visualizes how tokenization changes with vocabulary size on the first 32 bytes of the ELF header: the 64K tokenizer learns to merge the complete ELF64LE header prefix as a single token, demonstrating that larger vocabularies discover semantic file-format units. Table~\ref{tab:ls_anecdote} summarizes the corresponding compression values. Table~\ref{tab:bytes_per_token_vs_vocab} extends this evaluation across all tokenizers on 25 binaries (24 from the \texttt{samples/} benchmark plus \texttt{/usr/bin/ls}) spanning Linux, Windows, macOS, and Android platforms, showing that compression improves monotonically with vocabulary size across all operating systems and file formats.

\paragraph{32K vs.\ 64K comparison.} The 64K tokenizer wins every file, improving average bytes per token from 2.34 to 2.62 (an 11.7\% relative gain in compression efficiency). These improvements translate directly into larger effective context windows for downstream models, especially in settings where binaries and scripts co-occur.

\paragraph{Vocabulary relationship.} The entire \tokenizerfamilyname{} exhibits a perfect nested hierarchy: 4K $\subset$ 8K $\subset$ 16K $\subset$ 32K $\subset$ 64K. Every token ID in a smaller vocabulary maps to the identical byte sequence in all larger vocabularies, and all BPE merge operations occur in the same deterministic order. The first 4,089 tokens of the 4K tokenizer are identical to tokens 0--4088 in every larger tokenizer; for example, token ID 2048 encodes the identical 2-byte sequence (\texttt{0xe0a4}) across all five vocabularies. This nesting is a direct consequence of the BPE training algorithm: larger vocabularies simply continue merging from where smaller vocabularies stopped. The property enables seamless embedding transfer when scaling model capacity. A practitioner can train a 1B-parameter model with the 4K tokenizer (4K$\times$d embedding matrix), then scale to a 7B-parameter model with the 64K tokenizer by initializing the first 4K rows of the 64K$\times$d embedding matrix with the learned 4K embeddings and training only the 60K new token embeddings from scratch. This preserves all learned byte-sequence representations while avoiding costly retraining, supporting progressive model development from resource-constrained prototypes to production-scale deployments.

\begin{table}[t]
\centering
\footnotesize
\begin{tabular}{rrrp{1.8cm}|rrrp{1.8cm}}
\toprule
\multicolumn{4}{c|}{\textbf{4K Tokenizer}} & \multicolumn{4}{c}{\textbf{64K Tokenizer}} \\
\cmidrule(lr){1-4} \cmidrule(lr){5-8}
\textbf{Seq} & \textbf{Token ID} & \textbf{Size} & \textbf{Bytes} & \textbf{Seq} & \textbf{Token ID} & \textbf{Sz} & \textbf{Bytes} \\
\midrule
1 & 127 & 1 & \texttt{7F} & 1 & 45813 & 7 & \texttt{7F 45 ...} \\
2 & 3732 & 2 & \texttt{45 4C} & 2 & 662 & 9 & \texttt{00 00 ...} \\
3 & 70 & 1 & \texttt{46} & 3 & 265 & 2 & \texttt{03 00} \\
4 & 2 & 1 & \texttt{02} & 4 & 1369 & 2 & \texttt{3E 00} \\
5 & 392 & 2 & \texttt{01 01} & 5 & 38243 & 5 & \texttt{01 00 ...} \\
6 & 662 & 9 & \texttt{00 00 ...} & 6 & 74 & 1 & \texttt{4A} \\
7 & 265 & 2 & \texttt{03 00} & 7 & 1877 & 6 & \texttt{10 00 ...} \\
8 & 1369 & 2 & \texttt{3E 00} & — & — & — & — \\
9 & 279 & 4 & \texttt{01 00 ...} & — & — & — & — \\
10 & 240 & 1 & \texttt{F0} & — & — & — & — \\
11 & 74 & 1 & \texttt{4A} & — & — & — & — \\
12 & 1877 & 6 & \texttt{10 00 ...} & — & — & — & — \\
\bottomrule
\end{tabular}
\caption{Visual ablation: 4K vs 64K tokenization of the first 32 bytes of \texttt{/usr/bin/ls}. Each row shows one token. Columns: sequence number, token ID from vocabulary, size in bytes, first 2 bytes (or full content if $\leq$2). The 64K tokenizer learns multi-byte tokens that capture file format structure, while 4K fragments the same content into smaller atomic units.}
\label{tab:ls_visual_ablation}
\end{table}

\begin{table}[t]
\centering
\begin{tabular}{lrrrrr}
\toprule
\textbf{Vocab} & \textbf{4K} & \textbf{8K} & \textbf{16K} & \textbf{32K} & \textbf{64K} \\
\midrule
\textbf{Bytes/Token} & 1.773 & 1.931 & 2.106 & 2.315 & 2.534 \\
\bottomrule
\end{tabular}
\caption{Anecdote: bytes per token for \texttt{/usr/bin/ls} (measured).}
\label{tab:ls_anecdote}
\end{table}

\begin{table}[t]
\centering
\caption{Bytes per token vs. vocabulary size for the \tokenizerfamilyname{} across 25 evaluation binaries: Linux (16 samples), Windows (6), macOS (2), Android (1). Compression improves monotonically with vocabulary size across all platforms.}
\label{tab:bytes_per_token_vs_vocab}
\begin{tabular}{lrrrrr}
\toprule
 \textbf{Platform}  & \textbf{4K} & \textbf{8K} & \textbf{16K} & \textbf{32K} & \textbf{64K} \\
\midrule
 Linux & 1.625 & 1.769 & 1.937 & 2.139 & 2.385 \\
 Windows & 2.119 & 2.341 & 2.595 & 2.905 & 3.274 \\
 macOS & 2.183 & 2.333 & 2.542 & 2.806 & 3.058 \\
 Android & 1.074 & 1.102 & 1.152 & 1.242 & 1.411 \\
 \textbf{Overall} & \textbf{1.766} & \textbf{1.925} & \textbf{2.112} & \textbf{2.341} & \textbf{2.613} \\
\bottomrule
\end{tabular}
\end{table}

\begin{figure}[t]
	\centering
	\includegraphics[width=0.8\linewidth]{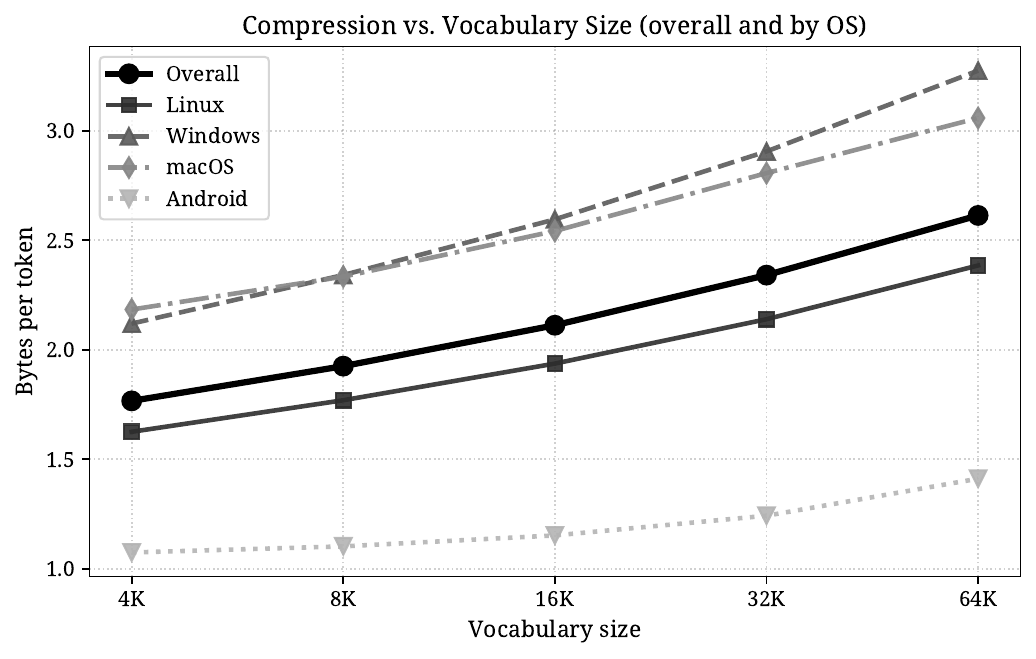}
	\caption{Bytes per token vs.\ vocabulary size for the \tokenizerfamilyname{} on the 25-file evaluation suite (24 binaries from the stratified \texttt{samples/} benchmark plus \texttt{/usr/bin/ls}), showing overall (thick line) and per-OS averages. Compression improves steadily with vocabulary size across all platforms.}
	\label{fig:scaling_curve}
\end{figure}

\begin{figure}[t]
	\centering
	\includegraphics[width=0.8\linewidth]{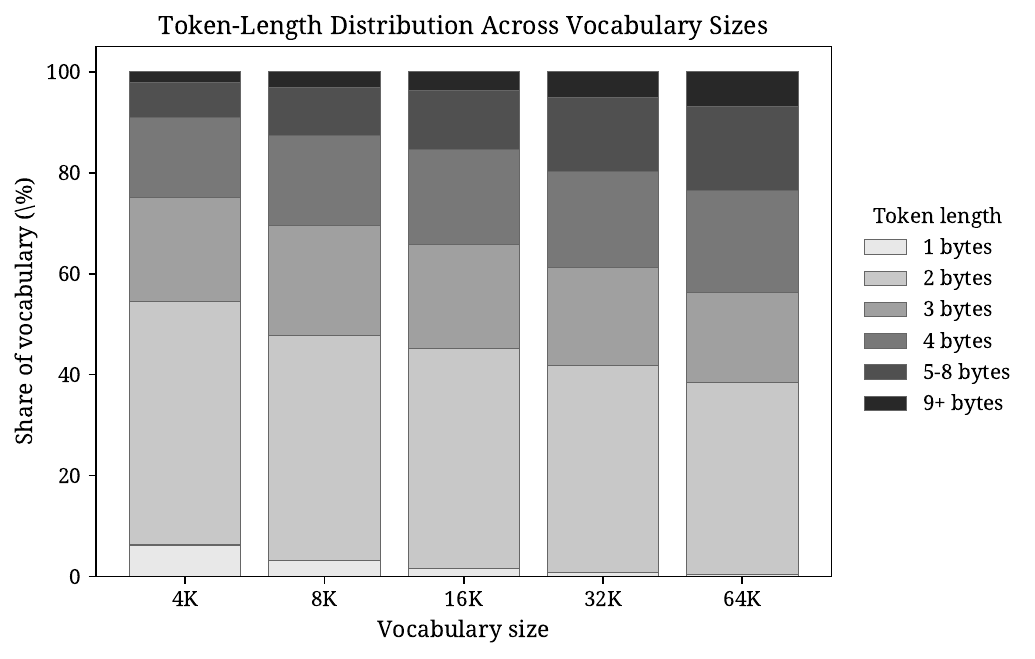}
	\caption{Token-length distributions (stacked) across vocabulary sizes.}
	\label{fig:token_length_stacks}
\end{figure}

\begin{figure}[t]
	\centering
	\includegraphics[width=0.8\linewidth]{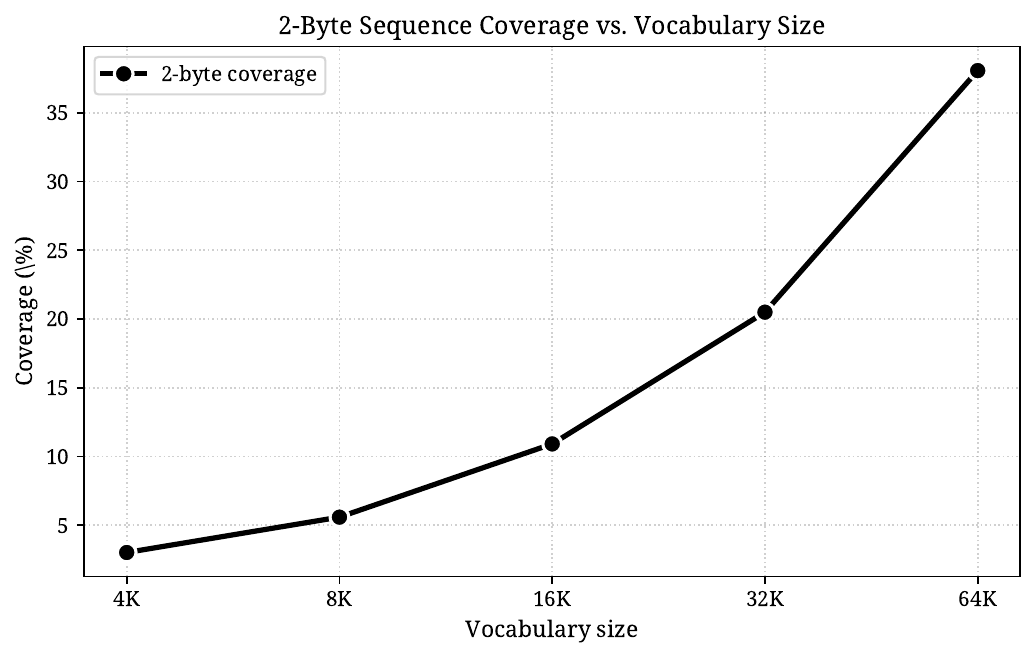}
	\caption{Coverage of 2-byte sequences vs.\ vocabulary size.}
	\label{fig:coverage_vs_vocab}
\end{figure}

These scaling results demonstrate that vocabulary size directly controls the compression--expressiveness tradeoff, with the 64K tokenizer unsurprisingly achieving the strongest performance across diverse binary types. Consistent improvements across platforms when moving from 32K to 64K confirm that larger vocabularies capture more sophisticated multi-byte patterns without sacrificing cross-platform generalization. In Section~\ref{sec:discussion}, we position this approach against alternative binary tokenization methods, provide practical deployment guidance for different vocabulary sizes, and discuss limitations and future directions.

\section{Discussion}
\label{sec:discussion}

\subsection{Positioning Against Alternative Approaches}

We compare our approach against alternatives for binary tokenization across four dimensions: compression efficiency, semantic preservation, robustness to adversarial cases, and cross-platform applicability.

\textbf{Raw bytes:} Operating directly on bytes~\cite{lu2025progressive} achieves 1.0 bytes/token (no compression), maximizing information preservation but wasting context. A 32KB binary consumes 32,768 tokens, leaving minimal room for analysis in typical context windows. In contrast, the 64K tokenizer typically achieves between roughly 2.5 and 3.5~bytes/token on realistic, predominantly uncompressed binaries (Section~\ref{sec:results}), with compressed APKs closer to 2.0~bytes/token, allowing models to process approximately 2--3$\times$ more binary content per forward pass on most platforms---critical for both model training and real-world deployment.

\textbf{Fixed n-grams:} Using fixed 2-byte, 4-byte, or 8-byte units provides deterministic compression (2.0, 4.0, or 8.0 bytes/token) without training. However, fixed boundaries fragment natural patterns: an x86-64 instruction spanning 3 bytes becomes split across tokens, preventing models from learning instruction-level semantics. Our adaptive BPE approach discovers natural boundaries, regardless of their size, and achieves stronger compression than fixed 2-grams while maintaining semantic coherence impossible with rigid segmentation.

\textbf{Assembly-level tokenization:} DisASM~\cite{mostafa2025tokenization} operates on disassembled instructions, providing semantic granularity (opcodes, operands, symbols) and potentially higher compression if assembly text is shorter than raw bytes. However, this requires accurate disassembly, which may be impacted by binary stripping, obfuscation, or section layouts. Our byte-level approach sacrifices symbolic annotations but handles any binary format robustly, critical for most production deployments.

\textbf{Domain-specific tokenizers:} MalBERTv2~\cite{rahali2023malbertv2} trains on Android APKs specifically, achieving specialized vocabulary for DEX bytecode and manifest structures. While this maximizes performance on a single platform, it requires separate tokenizers for each domain (Windows PE, Linux ELF, etc.). Our cross-platform tokenizer achieves consistent compression across platforms on both the \texttt{samples/} benchmark and the larger Binary-30K corpus (Section~\ref{sec:results}), enabling unified analysis pipelines and transfer learning across wider environments. Researchers studying cross-platform malware or library portability benefit from a single vocabulary capturing shared byte patterns alongside platform-specific structures.

Our approach occupies an optimal point in the design space: up to roughly 3$\times$ more efficient than raw bytes on typical uncompressed binaries, more robust than assembly-level methods requiring disassembly, and more generalizable than domain-specific tokenizers while maintaining competitive compression.

These trade-offs inform practical deployment decisions. The \tokenizerfamilyname's range of vocabulary sizes enables tailored configurations for different resource constraints.

\subsection{Practical Guidance by Vocabulary Size}

The \tokenizerfamilyname enables deployment across diverse resource constraints—from edge devices to datacenters—while supporting controlled ablation studies of vocabulary-size scaling behavior on binary data.

\paragraph{Resource-constrained environments.}
\textbf{4K/8K vocabularies} suit edge devices, IoT gateways, and embedded malware scanners where memory footprint and inference latency are critical. These configurations offer moderate compression while keeping the vocabulary footprint extremely small (tens of kilobytes on disk), enabling on-device binary classification in network security appliances performing inline packet inspection or firmware analysis in resource-limited industrial control systems. The small vocabulary reduces embedding layer size proportionally, enabling deployment on hardware accelerators with limited memory.

\paragraph{Balanced deployments.}
\textbf{16K/32K vocabularies} provide strong compression-resource tradeoffs for cloud-based malware analysis platforms and research prototypes. As shown in the ablation study (Section~\ref{sec:scaling}), these sizes capture most of the available compression gains on benchmarks like \texttt{/usr/bin/ls} and mixed binary workloads while keeping embedding layers compact. They are sufficient for most binary analysis tasks and suit exploratory research where model iteration speed matters more than maximum compression.

\paragraph{High-throughput datacenters.}
\textbf{64K vocabulary} maximizes compression and coverage for large-scale binary processing where accelerator memory and throughput dominate costs. The ablation study (Section~\ref{sec:scaling}) shows that the 64K tokenizer consistently outperforms 32K across a diverse evaluation suite, improving bytes per token by roughly 10--15\% while preserving backwards compatibility with the 32K vocabulary. The vocabulary size increase (64K vs.\ 32K) is negligible compared to typical transformer parameter counts (10M--1B+), making this the preferred choice when computational resources permit.

\subsection{Downstream Applications and Future Work}

This tokenizer enables a research agenda in transformer-based binary analysis and, more broadly, sequence-model-based binary tooling. In companion work we have used it to create the Binary-30K dataset~\cite{bommarito2025binary30k}, a large collection of pre-tokenized binaries with rich metadata for downstream task development; that paper provides the detailed dataset description. Using Binary-30K and the \tokenizerfamilyname{}, we are training transformer models for benchmark tasks including malware family classification, binary similarity detection, and function-purpose identification, with the goal of releasing public baselines for the community. Longer term, we envision smaller specialist masked- and causal-language models trained on the \tokenizerfamilyname{} and embedded as tools inside reverse-engineering frameworks (e.g., Ghidra, IDA Pro) or orchestrated by larger agentic systems and human analysts to provide interactive assistance with disassembly, decompilation, and optimization. For optimization tasks, the learned vocabulary enables detection of inefficient instruction sequences, dead code patterns, and redundant register operations---allowing both compiler-assisted code improvement and human-driven refactoring of handwritten or legacy binaries.

\subsection{Limitations and Boundary Conditions}

While our tokenizer achieves effective cross-platform compression, several limitations define its applicability boundaries and motivate future work.

\textbf{Adversarial robustness.} Packed and obfuscated malware pushes tokenization toward the theoretical lower bound for random data, where each token covers only a small number of bytes. Adversaries aware of tokenization-based analysis could deliberately inject high-entropy padding or employ polymorphic encryption to degrade compression, forcing models to waste context on near-random bytes. This represents a fundamental limitation of any compression-based approach: high entropy is incompressible. Detection of packing stubs and selective tokenization of executable sections could partially mitigate this, but an adaptive adversary can always maximize entropy.

A second class of limitations stems from data characteristics: binaries with extreme properties push against our tokenizer's assumptions and training distribution.

\textbf{Extreme sequence lengths.} Aggregate statistics on Binary-30K (Table~\ref{tab:large_corpus_stats}) show that Android APKs typically tokenize to well over a million tokens per file, exceeding most transformer context windows (8K--128K tokens) even after compression. Hierarchical tokenization (applying BPE recursively at different granularities) or selective windowing (identifying high-importance sections via entropy or disassembly) could address this, but introduce architectural complexity and potential information loss.  This example highlights the need for more sophisticated systems that orchestrate the unpacking or deobfuscation of binaries before tokenization and subsequent analysis.

\textbf{Architecture imbalance.} Less common architectures such as MIPS and RISC-V are underrepresented in the training corpus compared to x86-64 and ARM. While platform stratification prevents complete neglect, the vocabulary inevitably reflects corpus statistics, and pilot experiments suggest that underrepresented architectures achieve somewhat weaker compression. Training domain-specific tokenizers for specialized architectures or employing vocabulary augmentation techniques from low-resource NLP could improve coverage, at the cost of maintaining multiple tokenizers.

Beyond data challenges, the tokenizer faces a fundamental semantic limitation: its learned patterns capture byte statistics but lack execution-level grounding.

\textbf{Semantic gap.} BPE discovers byte co-occurrence patterns, not execution semantics. A token representing \texttt{0x48 0x89 0xE5} captures a frequent byte sequence (x86-64 MOV instruction) but lacks semantic grounding—the tokenizer does not know this moves a stack pointer. Bridging this gap through semi-supervised learning (training with small labeled instruction sets) or multi-task objectives (predicting both tokens and instruction types) could create semantically-aware embeddings, enabling better downstream task performance.

Finally, practical deployment faces a temporal challenge that is endemic to tokenizer-based systems:

\textbf{Format evolution.} Binary file formats evolve: new PE sections, updated ELF extensions, novel Android manifest structures. A tokenizer trained in 2025 may exhibit degraded performance on 2030 binaries with new format conventions. Periodic retraining or online vocabulary adaptation could maintain effectiveness, but introduces deployment complexity for production systems requiring stability.

\section{Conclusion}
\label{sec:conclusion}

We have introduced the \tokenizerfamilyname, an openly available, cross-platform set of BPE tokenizers trained directly on raw executable bytes. Built on a 24~GB corpus of approximately 30{,}000 binaries (the Binary-30K dataset) spanning ELF, PE, Mach-O, APK, and malware sources, these tokenizers learn a shared vocabulary that respects both platform-specific structure and architecture-level regularities. Across this corpus, the 64K configuration achieves multi-byte compression per token (roughly 2--3~bytes/token on typical uncompressed binaries, with compressed APKs closer to 2~bytes/token), enabling transformers and other sequence models to process up to about 2--3$\times$ more binary content per fixed-length context window than byte-level baselines while still exposing interpretable units such as file-format headers, instruction sequences, and cross-platform strings.

Our scaling study shows that growing the vocabulary from 4K to 64K steadily increases average token length and 2-byte sequence coverage, with the 64K tokenizer consistently dominating smaller variants on compression benchmarks while preserving a perfect nested hierarchy: each smaller vocabulary is a strict prefix of all larger ones. This design allows practitioners to scale models without discarding prior work---initializing larger embedding matrices from smaller configurations---and to choose among 4K/8K/16K/32K/64K vocabularies according to resource constraints, from embedded devices to high-throughput datacenters. Together with detailed vocabulary analysis and compression measurements on both stratified samples and the Binary-30K corpus, the Binary BPE family provides a practical, empirically grounded answer to the tokenization bottleneck in binary analysis.

To make these tokenizers usable beyond this paper, we release all \tokenizerfamilyname{} variants as HuggingFace-compatible models (\texttt{mjbommar/binary-tokenizer-001-\{4k,8k,16k,32k,64k\}}), so they can be dropped into existing transformer stacks and binary-analysis pipelines without retraining. Companion work leverages the same tokenizers to construct the Binary-30K dataset and to train transformer baselines for malware classification, binary similarity, and function-purpose identification, and we expect future research to build semantically-aware models and agentic tools on top of this shared byte-level vocabulary. By decoupling tokenizer design from model architecture and establishing a scalable, cross-platform foundation, the \tokenizerfamilyname{} aims to become a standard building block for future work in executable understanding, security analysis, and performance optimization.

\bibliographystyle{plain}
\bibliography{bibtex/references}

\end{document}